\documentclass[letterpaper, 10 pt, conference]{ieeeconf}  %

\IEEEoverridecommandlockouts                              %

\usepackage{graphics} %
\usepackage{epsfig} %
\usepackage{mathptmx} %
\usepackage{times} %
\usepackage{amsmath} %
\usepackage{amssymb}  %

\usepackage{bbm}
\usepackage{multirow}
\usepackage[table]{xcolor}
\usepackage{xcolor}

\usepackage{pifont}       %
\usepackage{bbding}       %
\usepackage{fontawesome}  %

\usepackage{booktabs}
\usepackage{hyperref}
\usepackage[capitalise]{cleveref}
\usepackage[table]{xcolor}

\title{\LARGE \bf
MambaMap: Online Vectorized HD Map Construction using

State Space Model
}

\author{Ruizi Yang$^{1}$, Xiaolu Liu$^{2}$, Junbo Chen$^{3}$, and Jianke Zhu$^{2}$%
\thanks{$^{1}$Ruizi Yang is with the College of Software Technology, Zhejiang University, Hangzhou 310027, China. { \tt\small ruiziyang@zju.edu.cn}%
}
\thanks{$^{2}$Xiaolu Liu and Jianke Zhu are with the College of Computer Science, Zhejiang University, Hangzhou 310027, China. Jianke Zhu is the corresponding author. {\tt\small \{xiaoluliu, jkzhu\}@zju.edu.cn}
}%
\thanks{$^{3}$Junbo Chen is with the Udeer.ai, Hangzhou 310000, China. { \tt\small junbo@udeer.ai}%
}
}

\begin{document}

\maketitle
\thispagestyle{empty}
\pagestyle{empty}

\begin{abstract}

High-definition (HD) maps are essential for autonomous driving, as they provide precise road information for downstream tasks. Recent advances highlight the potential of temporal modeling in addressing challenges like occlusions and extended perception range. However, existing methods either fail to fully exploit temporal information or incur substantial computational overhead in handling extended sequences. To tackle these challenges, we propose MambaMap, a novel framework that efficiently fuses long-range temporal features in the state space to construct online vectorized HD maps. Specifically, MambaMap incorporates a memory bank to store and utilize information from historical frames, dynamically updating BEV features and instance queries to improve robustness against noise and occlusions. Moreover, we introduce a gating mechanism in the state space, selectively integrating dependencies of map elements in high computational efficiency. In addition, we design innovative multi-directional and spatial-temporal scanning strategies to enhance feature extraction at both BEV and instance levels. These strategies significantly boost the prediction accuracy of our approach while ensuring robust temporal consistency. Extensive experiments on the nuScenes and Argoverse2 datasets demonstrate that our proposed MambaMap approach outperforms state-of-the-art methods across various splits and perception ranges. Source code will be
available at https://github.com/ZiziAmy/MambaMap.
\end{abstract}
    
\section{Introduction}
\label{sec:intro}

High-definition (HD) maps play a crucial role in autonomous driving systems by providing centimeter-level road details to support downstream tasks, such as trajectory prediction~\cite{zhao2021tnt} and path planning~\cite{gao2020vectornet}. Traditional HD map construction pipelines primarily rely on offline SLAM-based methods~\cite{zhang2014loam,shan2018lego}, which are resource-intensive and challenging to update. In contrast, online vectorized HD map construction has emerged as a promising alternative, enabling real-time updates while reducing dependence on offline processing. Several approaches~\cite{li2022hdmapnet,liu2023vectormapnet,liao2022maptr,qiao2023end} have been proposed to enhance online map construction. These methods typically transform surround-view images into Bird's Eye View (BEV) representations and employ transformer-based decoders~\cite{vaswani2017attention} to predict map elements.
\begin{figure}[ht]
\raggedright
\setlength{\abovecaptionskip}{-0.1cm}
\includegraphics[width=1.0\linewidth]{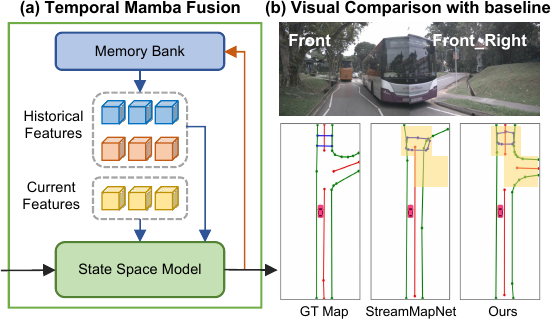}

\caption{(a) MambaMap framework for online HD map construction with efficient temporal fusion via a memory bank and state space model. (b) Compared to the baseline approach~\cite{yuan2024streammapnet}, improved map construction results can be obtained by MambaMap in complex occlusion scenarios.}
\label{fig:1}
   \vspace{-7mm}
\end{figure}

Although encouraging progress has been made, a major  challenge in vectorized HD map construction lies in handling occlusions from dynamic objects and maintaining accuracy over extended perception ranges. 
Recent studies~\cite{li2022bevformer,lin2023sparse4d,huang2022bevdet4d} show that temporal modeling mitigates these issues by leveraging sequential information to compensate for occlusions with historical context and improve long-range accuracy.
StreamMapNet~\cite{yuan2024streammapnet} recurrently propagates a single hidden state, offering lower latency than stacking-based methods~\cite{huang2022bevdet4d,yang2023bevformerv2} for online HD map construction.
 Several works~\cite{wang2024stream, chen2025maptracker, song2024memfusionmap} further introduce auxiliary denoising tasks, redundant memory buffers, and overlap information to enhance temporal modeling.

Despite these advances, we identify two critical limitations in existing temporal modeling approaches. Firstly, it is insufficient for capturing long-range temporal dependencies by relying solely on a single recurrent feature~\cite{chen2025maptracker,zhou2024rmem}. Secondly, maintaining and processing highly redundant sequences~\cite{chen2025maptracker} incurs substantial memory and computational costs, which poses challenges for real-time applications. 
To overcome these limitations, we propose to leverage the State Space Model (SSM)~\cite{gu2021combining,gu2021efficiently,gu2023mamba} that excels at capturing long-range dependencies with linear computational complexity. This makes SSM significantly more efficient than transformer-based temporal fusion methods~\cite{li2024bevformer,luo2022detr4d}, which suffer from quadratic complexity with respect to sequence length. 
Additionally, storage of redundant information can be alleviated by constraining the memory bank to a small set of highly relevant frames, allowing the model to focus more effectively on the most informative temporal features.

Building on these insights, we propose MambaMap (Fig.~\ref{fig:1}), a novel architecture designed for efficient temporal fusion in online vectorized HD map construction. To manage temporal information effectively, we employ a sliding-window-based memory bank that dynamically updates BEV features and instance queries at each time step. To achieve enhanced temporal modeling while maintaining high computational efficiency, we introduce a gating SSM mechanism that selectively integrates long-range dependencies, ensuring the model captures the most relevant temporal information. Furthermore, we design innovative scanning strategies at both the BEV and instance query levels, which incorporate multi-directional and spatial-temporal scanning mechanisms to fully exploit temporal features within the state space. These enhancements significantly improve the temporal consistency and accuracy of road element predictions.

In summary, our main contributions are as follows.
\begin{itemize}

\item A novel MambaMap framework that leverages SSM to efficiently fuse long-range temporal
information for online vectorized HD map construction.

\item 
An effective gating mechanism in state space for efficient information selection and integration at both the BEV feature and instance query levels. Moreover, distinct scanning strategies are designed to effectively exploit spatial-temporal dependencies.
\item Extensive experiments are conducted on the nuScenes and Argoverse2 datasets. Results across various settings consistently demonstrate the superiority of our approach compared to the state-of-the-art methods.

\end{itemize}

\section{Related Work}
\label{sec:related-work}

\noindent\textbf{Online Vectorized HD Map Construction.} 
Online vectorized HD map construction has received significant attention in recent years due to its high efficiency for real-time autonomous driving systems~\cite{li2022hdmapnet,liao2022maptr,liu2025uncertainty}.
HDMapNet~\cite{li2022hdmapnet} pioneers a post-processing approach to convert rasterized maps into vector maps, enhancing downstream usability. VectorMapNet~\cite{liu2023vectormapnet} proposes an end-to-end approach that makes use of an auto-regressive decoder to predict the points of map elements. MapTR~\cite{liao2022maptr} introduces a DETR-based~\cite{carion2020end} one-stage framework that utilizes hierarchical queries to predict all the vertices simultaneously. MapTRv2~\cite{liao2024maptrv2} further improves the decoder structure and introduces auxiliary tasks to enhance performance. Some other works explore diverse representations of map elements. BeMapNet~\cite{qiao2023end} represents map elements using Bézier curves, while PivotNet~\cite{ding2023pivotnet} adopts a pivot-based representation.

Temporal information has been proven highly effective in perception tasks, particularly in multi-view 3D detection and BEV feature learning~\cite{li2022bevformer,lin2023sparse4d,huang2022bevdet4d}. 
StreamMapNet~\cite{yuan2024streammapnet} integrates temporal cues into HD map construction via a streaming strategy by leveraging BEV feature fusion and query propagation. Later, SQD-MapNet~\cite{wang2024stream} enhances temporal consistency through a query denoising strategy inspired by DN-DETR~\cite{li2022dn}.  MapTracker~\cite{chen2025maptracker} reformulates map element detection as a tracking problem, maintaining historical information across 20 frames to exploit temporal dependencies. Recently, MemFusion~\cite{song2024memfusionmap} introduces overlap heatmaps to provide additional cues for improved temporal modeling.

\noindent\textbf{State Space Models (SSMs).} While transformers~\cite{vaswani2017attention,carion2020end} have been widely adopted in deep learning, their quadratic complexity poses significant challenges for processing long sequences. As a compelling alternative for modeling long-range dependencies, SSMs have garnered substantial attention due to their linear complexity with respect to sequence length. Gu~\textit{et al.}~\cite{gu2021combining} introduce the linear state-space layer, which combines recurrent neural networks with temporal convolutions to capture sequential patterns. To improve computational efficiency, S4~\cite{gu2021efficiently} employs a parameterized state matrix to compute outputs directly.
Subsequent explorations of SSMs include DSS~\cite{gupta2022diagonal} and GSS~\cite{mehta2022long}. DSS assumes a diagonal state matrix to simplify complexity. On the other hand, GSS incorporates gating mechanisms to capture complex dependencies. Recently, Mamba~\cite{gu2023mamba} introduces selective state spaces, which significantly enhances performance by effectively modeling diverse dependencies.

SSMs have also been extended to visual applications. To process images with linear complexity, Vision Mamba~\cite{zhu2024vision} and VMamba~\cite{liu2025vmamba} employ bidirectional and four-way scanning mechanisms, respectively. Video Mamba~\cite{li2025videomamba} incorporates dynamic spatial-temporal context modeling for efficient video understanding. Building on these advancements, we propose a state-space-based temporal fusion framework that efficiently integrates multi-frame features for online HD map construction.

\begin{figure*}[t!]
    \centering
    \vspace{0.1cm}
\setlength{\abovecaptionskip}{-0.25cm}
    \includegraphics[width=1.0\linewidth]{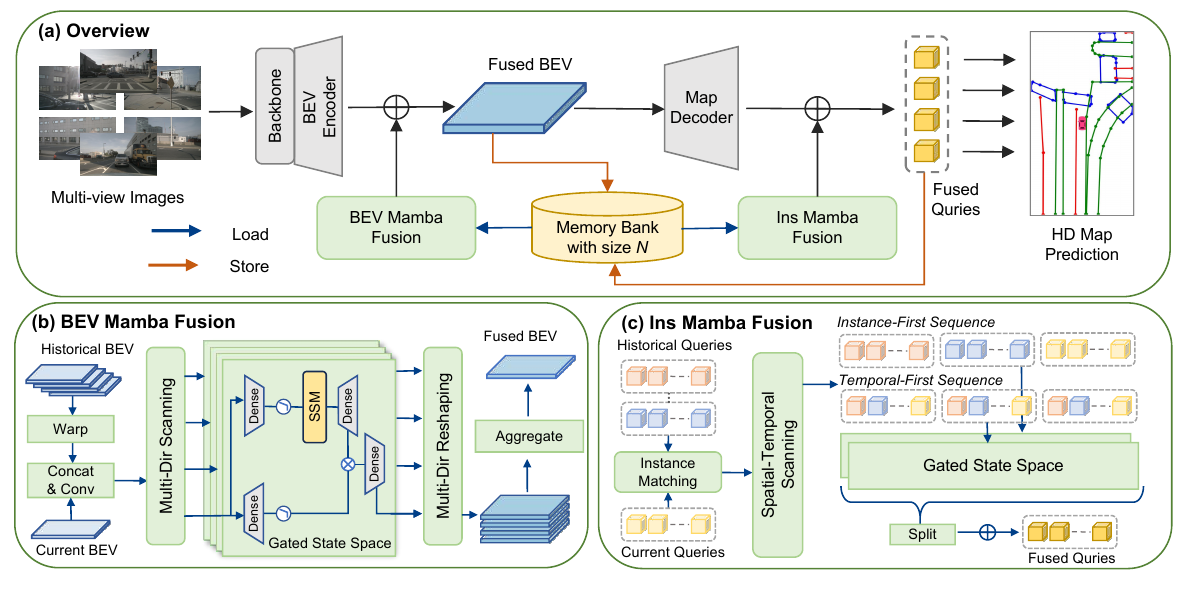}
  
    \caption{(a) Overview of our proposed MambaMap approach. Temporal fusion is performed at both BEV and instance levels using a Memory Bank for efficient temporal information management. (b) BEV Mamba Fusion. Multi-frame BEV features are processed via Gated State Space blocks with multi-directional scanning to capture map structure dependencies. (c) Instance Mamba Fusion. Multi-frame instance queries are refined through spatial-temporal scanning and State Space modeling to improve the accuracy.}
    
    \label{fig:2}
     \vspace{-6mm}
\end{figure*}

\section{Methodology}
In this section, we present MambaMap, an efficient architecture for online vectorized HD map construction. %

\subsection{Overview}
In this paper, we present a novel online vectorized HD map construction framework MambaMap that incorporates advanced temporal fusion strategies using SSM, as illustrated in Fig.~\ref{fig:2}. Specifically, the whole pipeline firstly extracts features from multi-view images captured by onboard cameras through a backbone and Feature Pyramid Network (FPN)~\cite{lin2017feature}. These features are then transformed into an initial BEV representation via a BEV Encoder, forming the foundation for subsequent temporal modeling.

Our design enhances temporal modeling through three key components, including a Memory Bank, the BEV Mamba Fusion module, and the Instance Mamba Fusion module. The Memory Bank maintains temporal context by storing historical features, providing critical support for the fusion process. The BEV Mamba Fusion module enhances initial BEV features by incorporating historical BEV information, which resolves ambiguities in map element extraction at the current time step. The enhanced BEV features are then fed into a Deformable DETR-based Map Decoder~\cite{zhu2020deformable}, interacting with learnable map queries. Subsequently, the Instance Mamba Fusion module further refines these queries by extracting temporal dependencies in order to improve the accuracy of point coordinate and category predictions for each map instance.
\subsection{Memory Bank}

To efficiently manage temporal information, our architecture incorporates a Memory Bank that stores and updates historical frame data. The Memory Bank employs a size-restricted mechanism that retains only the most recent and relevant features. Such design allows the model to focus on key spatial information while discarding redundant or noisy data, thereby enhancing both computational efficiency and representation quality.

The Memory Bank consists of two components, including $\mathbf{M}^{BEV}$ for BEV features and $\mathbf{M}^{Ins}$ for instance queries. At time $t$, it is represented as $\mathbf{M}_t = \{\mathbf{M}_t^{BEV}, \mathbf{M}_t^{Ins}\}$. 

The BEV feature memory $\mathbf{M}_t^{BEV}$ stores the refined BEV features from previous frames, which is defined as below:
\begin{equation}
\mathbf{M}_t^{BEV} = \{\mathbf{F}'_{t-N}, \mathbf{F}'_{t-N+1}, \dots, \mathbf{F}'_{t-1}\},
\end{equation}
where $\mathbf{F}'_{t-k}$ represents the refined BEV features from time $t-k$. $N$ represents the capacity of the Memory Bank to determine how many historical features are retained. 

Likewise, the instance query memory $\mathbf{M}_t^{Ins}$ stores refined instance queries from previous frames, which is defined as:
\begin{equation}
\mathbf{M}_t^{Ins} = \{\mathbf{Q}'_{t-N}, \mathbf{Q}'_{t-N+1}, \dots, \mathbf{Q}'_{t-1}\},
\end{equation}
where $\mathbf{Q}'_{t-k}$ denotes the refined instance queries from time step $t-k$. 

Overall, these components form the foundation for fusing historical and current data so that BEV feature representation and map instance detection could be enhanced.

\subsection{BEV Mamba Fusion} \label{sec:BEV}
Given historical and current BEV features, BEV Mamba Fusion (BMF) seeks to capture temporally consistent map information to generate enhanced BEV features.
The module firstly warps all historical BEV features in \( \mathbf{M}_t^{BEV} \) to align with the current frame:
\begin{equation}
\begin{aligned}
\mathbf{\widetilde{M}}_t^{BEV} = \{\text{Warp}(\mathbf{F}'_{t-k}, \mathbf{T}_{t-k \to t})\} \,|\, k \in \{1, 2, \dots, N\}\}, 
\end{aligned}
\end{equation}
where \( \mathbf{T}_{t-k \to t} \) is the \( 4 \times 4 \) affine transformation of the vehicle coordinate frame from time step \( t-k \) to \( t \). Each \( \mathbf{F}'_{t-k} \in \mathbb{R}^{C \times H \times W} \), where \( C \) denotes the dimension of the feature channel. 
\( H \) and 
\( W \) represent the spatial dimensions of the BEV feature, respectively. 

The aligned historical features \(\mathbf{\widetilde{M}}_t^{BEV} \) are concatenated with the current BEV feature \( \mathbf{F}_t \) along the channel dimension. Then, the concatenated features are processed through a convolutional layer followed by layer normalization:
\begin{equation}
\begin{aligned}
\mathbf{F}_c &= \text{Concat}(\mathbf{\widetilde{M}}_t^{BEV},\mathbf{F}_t),\\
\mathbf{F}_f &= \text{LayerNorm}(\text{Conv}(\mathbf{F}_c)),
\end{aligned}
\end{equation}
where \( \mathbf{F}_f \) denotes the fused BEV feature for the current frame.

\noindent\textbf{Multi-Directional Scanning.}
To fully exploit the capability of SSM in capturing long-range dependencies, we perform multi-directional scanning of the feature map
\( \mathbf{F}_f \) along four distinct directions, including left, right, up, and down, as illustrated in Fig.~\ref{fig:3} (a). This process generates 1D sequences represented as below:
\begin{align}
\mathbf{S} = \{\mathbf{S}_i \,|\, i \in \{\text{left}, \text{right}, \text{up}, \text{down}\}\},
\end{align}
where $i$ represents the scanning direction. Each \( \mathbf{S}_i \in \mathbb{R}^{{S}_L \times C } \), with \( S_L = H \cdot W \), represents the flattened sequence length.

The proposed scanning strategy enables SSM to capture spatial relationships from multiple orientations, thereby enhancing its sensitivity to local geometric patterns. This approach is particularly effective for modeling elongated structures of roads that exhibit strong directional characteristics in the BEV representation.%

\noindent\textbf{Gated State Space.} Subsequently, the sequences in \(\mathbf{S}\) are processed in parallel through distinct Gated State Space blocks, an SSM-based approach inspired by~\cite{mehta2022long}. Each sequence
\(\mathbf{S}_i\) is firstly projected into a gating signal \(\mathbf{V}_i \in \mathbb{R}^{S_L \times \beta C}\), where \( \beta \)  acts as a dimensionality expansion factor:
\begin{equation}
\mathbf{V}_i = \text{GELU}(\mathbf{W}_v (\mathbf{S}_i)),
\end{equation}
where \(\mathbf{W}_v \in \mathbb{R}^{C \times \beta C}\) refers to the learnable parameters of a Linear layer. $\text{GELU}$ is the Gaussian error linear units~\cite{hendrycks2016gaussian}.

\(\mathbf{S}_i\) is projected into a
lower-dimensional space \(\mathbf{U}_i \in \mathbb{R}^{S_L \times \alpha C}\), where \( \alpha \) serves as a dimensionality reduction factor. The sequence is then enhanced by the state space layer to produce a contextualized representation:
\begin{equation}
\begin{aligned}
\mathbf{U}_i &= \text{GELU}(\mathbf{W}_u (\mathbf{S}_i)),  & 
\mathbf{Y}_i &= \text{DSS}(\mathbf{U}_i),  
\end{aligned}
\end{equation}
where  \(\mathbf{W}_u \in \mathbb{R}^{C \times \alpha C}\) is the learnable parameters of a Linear layer, and $\text{DSS}$ refers to the Diagonal State Spaces layer proposed by~\cite{gupta2022diagonal}. This refined representation is further modulated by \(\mathbf{V}_i \)
  through the gating mechanism, ensuring that only critical contextual information is retained:
  \begin{equation}
\begin{aligned}
\mathbf{U}_i' &=\mathbf{W}_y (\mathbf{Y}_i), &
\mathbf{O}_i &= \mathbf{W}_o (\mathbf{U}_i' \odot \mathbf{V}_i) + \mathbf{S}_i,
\end{aligned}
\end{equation}
where \(\mathbf{W}_y \in \mathbb{R}^{\alpha C \times \beta C}\) and \(\mathbf{W}_o \in \mathbb{R}^{\beta C \times C}\)  stand for the learnable linear layer parameters, 
and \( \odot \) denotes element-wise multiplication. 
With the above design, the block achieves high computational efficiency while selectively retaining contextual information, effectively capturing long-range dependencies in BEV sequences.

The processed sequences \( \mathbf{O}_i \)
  are then reshaped back into the BEV grid according to their respective scanning direction. Finally, the refined BEV feature \( \mathbf{F}'_{t} \) is obtained by averaging the reconstructed features across all directions, thereby integrating the advantages of multi-directional sequence processing for the enhanced feature representation.%

\subsection{Instance Mamba Fusion}
\noindent\textbf{Instance Matching.}
To further leverage instance-level temporal information, we introduce Instance Mamba Fusion (IMF). Since the same map element instances are consistently detected across consecutive frames with high probability, it is essential to maintain their order for effective temporal modeling.

Given the instance query sequence \( \mathbf{M}_t^{Ins} = \{\mathbf{Q}'_{t-N}, \mathbf{Q}'_{t-N+1}, \dots, \mathbf{Q}'_{t-1}\} \) from the past \( N \) time steps and the current queries \( \mathbf{Q}_t \in \mathbb{R}^{N_{\text{q}} \times D} \) obtained from the decoder, where \( N_{\text{q}} \) denotes the number of queries per frame and 
 \( D\) represents the feature dimension of each query, we perform instance matching to ensure consistent ordering between historical and current queries.

To simplify computation and avoid additional supervision, we measure instance similarity using the $L_2$ distance between query vectors. The matching cost $d(\cdot)$ is defined as follows:
\begin{align}
d(\mathbf{q}_n^{(t-k)}, \mathbf{q}_m) = \|\mathbf{q}_n^{(t-k)} - \mathbf{q}_m\|_2^2, 
\end{align}
where $\mathbf{q}_n^{(t-k)} \in \mathbf{Q}'_{t-k}$ represents a historical query, and $\mathbf{q}_m \in \mathbf{Q}_t$ denotes a current query.

We then employ the Hungarian algorithm to match instances based on their costs, generating a permutation \( \mathbf{\pi}_{t-k \to t}\) for each historical time step: 
\begin{align}
\pi_{t-k \to t} = \arg\min_{\pi \in 
 \Pi_{N_{\text{q}}}}\sum_{i=1}^{N_{\text{q}}} d(\mathbf{q}_n^{(t-k)}, \mathbf{q}_m),
\end{align}
where $\Pi_{N_{\text{q}}}$ denotes the set of all possible permutations over $N_{\text{q}}$ elements.

Using this permutation, we reorder the historical instance queries \( \mathbf{Q}'_{t-k} \) as follows:
\begin{align}
\mathbf{\widetilde{M}}_t^{Ins} = \{\text{Permute}(\mathbf{Q}'_{t-k}, \mathbf{\pi}_{t-k \to t}) \,|\, k \in \{1, 2, \dots, N\}\}.
\end{align}

\begin{figure}[t]
\raggedright
\setlength{\abovecaptionskip}{-0.1cm}
\includegraphics[width=1.0\linewidth]{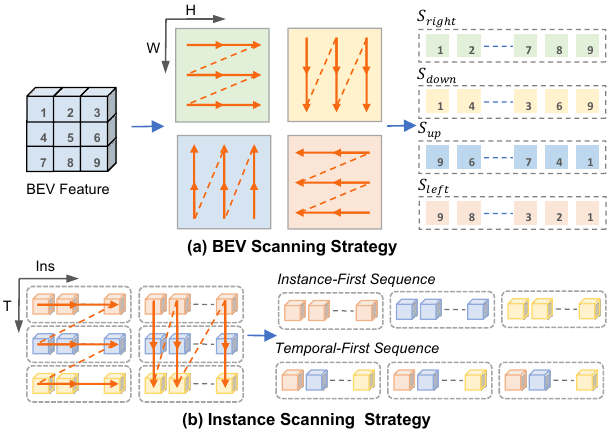}
\caption{(a) BEV Scanning Strategy: BEV features are processed using a multi-directional scanning mechanism to generate sequences with diverse patch arrangements. (b) Instance Scanning Strategy: Instance queries are organized into two distinct sequences, prioritizing either instance-centric or time-centric arrangements.}
\vspace{-6mm}
\label{fig:3}
\end{figure}

\noindent\textbf{Spatial-Temporal Scanning.}
Building upon the instance sequences \(\mathbf{\widetilde{M}}_t^{Ins} \) with consistent order, we introduce a novel spatial-temporal scanning mechanism based on SSM, as illustrated in Fig.~\ref{fig:3} (b). This method leverages two complementary sequence arrangements to capture both intra-frame interactions and inter-frame temporal dependencies. 

Firstly, we construct the \textit{Instance-First Sequence} $\mathbf{H}_s \in \mathbb{R}^{I_L \times D}$ by concatenating instances frame by frame in a front-to-back manner, where $I_L = (N+1) \cdot N_{\text{q}}$.
This arrangement emphasizes spatial relationships within each frame while capturing temporal evolution across frames, enabling the model to effectively learn intra-frame interactions and temporal dynamics.

Secondly, we create the \textit{Temporal-First Sequence} $\mathbf{H}_t \in \mathbb{R}^{T_L \times D}$ by stacking queries along the temporal dimension at the same time, where $T_L = N_{\text{q}} \cdot (N+1)$. This arrangement enhances temporal consistency of individual instances over time , allowing the model to focus on long-term dependencies and robustly track instance-specific features.

\begin{table*}[t]
  \footnotesize
  \centering
  \renewcommand\arraystretch{1.05}
  \renewcommand\tabcolsep{7pt}
  \vspace{0.2cm}
  \caption{Comparison with baseline methods on the original nuScenes split at both $30\,m$ and $50\,m$ perception ranges. Methods with “*” indicate results obtained from the corresponding papers, while others are reproduced using official public codes. ``EB0" and ``R50" denote the backbones Efficient-B0~\cite{tan2019efficientnet} and ResNet50~\cite{he2016deep}.}
  \vspace{-2mm}
  \label{tab:Nus}
  \resizebox{\textwidth}{!}{%
    \begin{tabular}{c|c|ccc|cccc|c}
      \toprule
      Range & Method & Backbone & Image Size & Epoch & AP$_{ped}$ & AP$_{div}$ & AP$_{bou.}$ & mAP & FPS \\
      \midrule
      \multirow{11}{*}{$60\times30\,m$}
        & HDMapNet~\cite{li2022hdmapnet}     & EB0 & $128\times352$ & 30  & 24.1 & 23.6 & 43.5 & 31.4 & – \\
        & VectorMapNet*~\cite{liu2023vectormapnet} & R50 & $256\times480$ & 110 & 36.1 & 47.3 & 39.3 & 40.9 & – \\
        & MapTR~\cite{liao2022maptr}         & R50 & $480\times800$ & 30  & 45.2 & 53.8 & 54.3 & 51.1 & \textbf{16.8} \\
        & GeMap~\cite{zhang2023gemap}        & R50 & $480\times800$ & 30  & 50.9 & 57.4 & 58.7 & 55.7 & 12.6 \\
        & PivotNet*~\cite{ding2023pivotnet}   & R50 & $512\times896$ & 30  & 53.8 & 58.8 & 59.6 & 57.4 & 11.3 \\
        & BeMapNet*~\cite{qiao2023end}        & R50 & $512\times896$ & 30  & 57.7 & 62.3 & 59.4 & 59.8 & 10.1 \\
        & MGMap~\cite{liu2024mgmap}          & R50 & $480\times800$ & 30  & 57.4 & 63.5 & 63.3 & 61.4 & 12.7 \\
        & MapTRv2~\cite{liao2024maptrv2}      & R50 & $480\times800$ & 30  & 61.9 & 64.1 & 64.0 & 63.3 & 14.5 \\
        & StreamMapNet~\cite{yuan2024streammapnet} & R50 & $480\times800$ & 30 & 60.9 & 66.3 & 63.0 & 63.4 & 14.9 \\
        & SQD-MapNet~\cite{wang2024stream}   & R50 & $480\times800$ & 30  & \underline{64.3} & \underline{67.1} & \underline{66.1} & \underline{65.8} & 14.4 \\
      \rowcolor{gray!15}
      \cellcolor{white} 
        & MambaMap (ours)                    & R50 & $480\times800$ & 30  & \textbf{66.6} & \textbf{68.0} & \textbf{67.2} & \textbf{67.3} & 12.7 \\
      \midrule
      \multirow{3}{*}{$100\times50\,m$}
        & StreamMapNet~\cite{yuan2024streammapnet} & R50 & $480\times800$ & 30 & 63.9 & \underline{66.3} & 60.4 & 63.5 & \textbf{14.9} \\
        & SQD-MapNet~\cite{wang2024stream}   & R50 & $480\times800$ & 30  & \underline{65.7} & 65.2 & \underline{61.5} & \underline{64.1} & 14.4 \\
      \rowcolor{gray!15}
      \cellcolor{white} 
        & MambaMap (ours)                    & R50 & $480\times800$ & 30  & \textbf{68.7} & \textbf{67.8} & \textbf{62.9} & \textbf{66.5} & 12.7 \\
      \bottomrule
    \end{tabular}%
  }
  \vspace{-7mm}
\end{table*}

Finally, both $\mathbf{H}_s$ and $\mathbf{H}_t$ are processed through the parallel Gated State Space blocks described in~\cref{sec:BEV}. These blocks efficiently model long-range dependencies and capture intricate spatial-temporal relationships among instances. After processing, the current frame's queries are split, aggregated into fused instance queries $\mathbf{Q}'_t$
 , and fed into classification and regression heads for final instance predictions.

\subsection{Loss Setting}
During training, our model performs Hungarian matching between predictions and ground-truth labels, followed by computing the loss for matched pairs as described in~\cite{yuan2024streammapnet}. The overall map loss $L_{\text{map}}$ is formulated as:
\begin{equation}
    L_{\text{map}} = \lambda_1 L_{\text{pts}} + \lambda_2 L_{\text{cls}},
\end{equation}
where $L_{\text{pts}}$ denotes the polyline-wise matching cost computed using $SmoothL_1$ loss, and $L_{\text{cls}}$ represents the classification cost computed using $Focal$ loss. $\lambda_1$ and $\lambda_2$ are hyperparameters that balance the contributions of the two loss terms.

To ensure comprehensive optimization, both the original queries from the decoder layer and the fused queries are supervised by the above loss functions.

\section{Experiments}
\label{sec:experiments}
\subsection{Experimental Settings}

\noindent\textbf{Datasets.} 
We conduct extensive experiments on two widely-used public datasets, nuScenes~\cite{nuScenes} and Argoverse2~\cite{Argoverse2}. nuScenes dataset consists of 1,000 scenes collected from Boston and Singapore, with annotations provided at a keyframe rate of 2 Hz, and each frame contains six surrounding-view images. Argoverse2 comprises 1,000 scenes from six cities, with images collected by seven cameras per scene at a 10 Hz annotation rate. In our experiments, the sampling frequency of Argoverse2 is unified to 2 Hz, which is consistent with the one in~\cite{yuan2024streammapnet}.

\noindent\textbf{Evaluation Metrics.}
To ensure fair comparisons, we focus on three map elements, including lane-divider ($div.$), pedestrian-crossing ($ped.$), and road-boundary ($bou.$). Performance is evaluated over two perception ranges, a small range (60 $\times$ 30m) and a larger range (100 $\times$ 50m). We adopt Average Precision (AP) as the evaluation metric, which is calculated using distance thresholds of \{0.5m, 1.0m, 1.5m\} for the small range and \{1.0m, 1.5m, 2.0m\} for the larger range. The mean Average Precision (mAP) is obtained by averaging AP across all three map types.

\noindent\textbf{Implementation Details.}
Our model is trained on two NVIDIA A30 GPUs with a batch size of 8. We use the AdamW~\cite{loshchilov2017fixing} optimizer at a learning rate of $2.5 \times 10^{-4}$.  We employ ResNet-50~\cite{he2016deep} for image feature extraction, while BEVFormer~\cite{li2024bevformer} serves as the BEV feature encoder. We set $N = 4$, $H = 50$, $W = 100$, $N_{\text{num}} = 100$, $D = 512$, $\alpha = 0.5$, $\beta = 4$, $\lambda_1 = 5.0$ and $\lambda_2 = 50.0$ as the hyperparameters across all experiments. Training follows a two-stage procedure~\cite{yuan2024streammapnet}. In the first stage, single-frame inputs are used to stabilize training. In the second stage, consecutive sequences are randomly split into two parts. Memory bank features are detached to prevent gradient propagation to historical frames.

\subsection{Comparisons with State-of-the-arts}
\noindent\textbf{Evaluation on nuScenes.}
 We compare our proposed MambaMap with previous methods on the nuScenes validation set with two perception ranges. Inference speed is measured on a single NVIDIA A30 GPU.
 
 As shown in Table~\ref{tab:Nus}, MambaMap achieves 67.3 mAP and 66.5 mAP for the small and large range, respectively. Compared to the baseline approach StreamMapNet, our model demonstrates significant improvements with 3.9 mAP and 3.0 mAP. Our proposed MambaMap consistently outperforms all other methods at substantial margins without relying on additional supervision, such as semantic segmentation or temporal correspondence between ground truth annotations.
 
\begin{table*}[t]
  \centering
  \renewcommand\arraystretch{1.05}
  \renewcommand\tabcolsep{7pt}
  \footnotesize
  \caption{Comparison with baseline methods on the original Argoverse2 split at both $30\,m$ and $50\,m$ perception ranges. Results marked with “*” are from the original papers. ``R50" corresponds to the backbone ResNet50~\cite{he2016deep}.}
  \label{tab:AV2}
  \vspace{-2mm}
  \resizebox{\textwidth}{!}{%
    \begin{tabular}{c|c|ccc|cccc|c}
    \toprule
    Range & Method & Backbone & Image Size & Epoch & AP$_{ped}$ & AP$_{div}$ & AP$_{bou.}$ & mAP & FPS \\
    \midrule
    \multirow{5}{*}{$60\times30\,m$}
      & VectorMapNet*~\cite{liu2023vectormapnet} & R50 & – & 110 & 38.3 & 36.1 & 39.2 & 37.9 & – \\
      & MapTR~\cite{liao2022maptr}              & R50 & $608\times608$ & 30 & 50.6 & 60.7 & 61.2 & 57.4 & \textbf{17.4} \\
      & StreamMapNet~\cite{yuan2024streammapnet}& R50 & $608\times608$ & 30 & 61.2 & 60.0 & 62.1 & 61.1 & 15.9 \\
      & SQD-MapNet*~\cite{wang2024stream}       & R50 & $608\times608$ & 30 & \underline{64.9} & \underline{60.2} & \underline{64.9} & \underline{63.3} & 15.4 \\
    \rowcolor{gray!15}
    \cellcolor{white} 
      & MambaMap (ours)                         & R50 & $608\times608$ & 30 & \textbf{65.4} & \textbf{61.1} & \textbf{68.3} & \textbf{64.9} & 13.6 \\
    \midrule
    \multirow{3}{*}{$100\times50\,m$}
      & StreamMapNet~\cite{yuan2024streammapnet}& R50 & $608\times608$ & 30 & 64.1 & \underline{55.9} & 53.6 & 57.9 & \textbf{15.9} \\
      & SQD-MapNet*~\cite{wang2024stream}       & R50 & $608\times608$ & 30 & \underline{66.9} & 54.9 & \underline{56.1} & \underline{59.3} & 15.4 \\
    \rowcolor{gray!15}
    \cellcolor{white} 
      & MambaMap (ours)                         & R50 & $608\times608$ & 30 & \textbf{68.6} & \textbf{56.1} & \textbf{57.5} & \textbf{60.7} & 13.6 \\
    \bottomrule
    \end{tabular}%
  }
  \vspace{-6mm}
\end{table*}

Considering the location overlaps between the original training and validation sets, we further evaluate our model on the geographically disjoint split of the nuScenes dataset proposed in~\cite{yuan2024streammapnet}. The new split ensures that the training and validation sets are strictly separated by geography, providing a more rigorous assessment of generalization performance. Models are trained for 24 epochs in the new split like~\cite{yuan2024streammapnet,song2024memfusionmap}. 
As shown in Table~\ref{tab:Newsplit}, our MambaMap outperforms all competing methods with a minimum gain of 3.1 mAP. These promising results demonstrate the strong generalization ability of MambaMap in dealing with unseen geographical locations.

\begin{table}[t]
  \centering
  \vspace{0.12cm}
  \renewcommand\arraystretch{1.05}
  \renewcommand\tabcolsep{2pt}
  \footnotesize
  \caption{Performance comparison on the new split of nuScenes and Argoverse2 at $30\,m$ range. Results with “*” are from the original papers.}
  \label{tab:Newsplit}
  \vspace{-2mm}
  \resizebox{\columnwidth}{!}{%
    \begin{tabular}{c|c|cccc}
      \toprule
      Dataset & Method & AP$_{ped}$ & AP$_{div}$ & AP$_{bou.}$ & mAP \\
      \midrule
      \multirow{4}{*}{nuScenes}
        & StreamMapNet*~\cite{yuan2024streammapnet} & 32.2 & 29.3 & 40.8 & 34.1 \\
        & SQD-MapNet~\cite{wang2024stream}          & 37.4 & 31.1 & 43.5 & 37.3 \\
        & MemFusionMap*~\cite{song2024memfusionmap} & \underline{38.3} & \underline{32.1} & \underline{43.6} & \underline{38.0} \\
      \rowcolor{gray!15}
      \cellcolor{white} 
        & MambaMap (ours)                           & \textbf{40.2} & \textbf{33.4} & \textbf{46.6} & \textbf{40.1} \\
      \midrule
      \multirow{4}{*}{Argoverse2}
        & StreamMapNet*~\cite{yuan2024streammapnet} & 56.9 & 55.9 & 61.4 & 58.1 \\
        & SQD-MapNet~\cite{wang2024stream}          & \textbf{59.4} & \underline{57.8} & 64.7 & \underline{60.6} \\
        & MemFusionMap*~\cite{song2024memfusionmap} & \underline{59.3} & 57.2 & \underline{65.1} & \underline{60.6} \\
      \rowcolor{gray!15}
      \cellcolor{white} 
        & MambaMap (ours)                           & \textbf{59.4} & \textbf{58.4} & \textbf{65.2} & \textbf{61.0} \\
      \bottomrule
    \end{tabular}%
  }
  \vspace{-6mm}
\end{table}

\noindent\textbf{Evaluation on Argoverse2.} 
More experiments are conducted on Argoverse2 dataset. For a fair comparison, all models are trained for 30 epochs. As shown in Table~\ref{tab:AV2}, our proposed MambaMap achieves 64.9 mAP under the small perception range and 60.7 mAP under the large perception range, which outperforms the previous approaches. Moreover, Table~\ref{tab:Newsplit} presents our results on the geographically disjoint split of Argoverse2. Our method achieves the highest mAP across all evaluation metrics, demonstrating the consistent effectiveness of our proposed method.

\subsection{Ablation Study}
We conduct ablation studies to evaluate the effectiveness of our proposed modules and designs. All experiments are performed on the original nuScenes split with a perception range of $60 \times 30m$, training for 30 epochs.

Table~\ref{tab:Fusion} evaluates the impact of Temporal Fusion Modules at different levels, using StreamMapNet without the streaming strategy as the baseline. Introducing the BMF to extract multi-frame road BEV features improves mAP by 3.7. Adding the IMF to refine instance queries boosts mAP by 2.2 over the baseline. Our method achieves the highest performance with 67.3 mAP by integrating both modules.%

We also evaluate the impact of different scanning strategies, as shown in Table~\ref{tab:Scanning}. For BEV Fusion, single-directional scanning achieves an mAP of 64.9, while bidirectional horizontal and vertical scanning improve this to 66.1 and 66.3, respectively. Multi-directional scanning, which integrates information from all directions, achieves the best mAP of 67.3. At the instance level, spatial and temporal scanning strategies are applied separately, whose combination yields the highest performance.
\begin{table}[t]
\vspace{0.12cm}
\centering

  \renewcommand\arraystretch{1.05}
  \renewcommand\tabcolsep{9pt}
\footnotesize
\caption{Ablation Study of Temporal Fusion Module at BEV and Instance Levels.}
   \vspace{-2mm}
\resizebox{\columnwidth}{!}{%
  \begin{tabular}{ cc|cc c c}
    \toprule
BMF & IMF & AP$_{\textit{ped}}$ & AP$_{\textit{div}}$ & AP$_{\textit{bou}}$ &  mAP \\ 
    \midrule
     &   & 59.4  &64.2 &60.9 & 61.5   \\
     \Checkmark &  &65.5   &65.6  &64.6 &65.2      \\ 
     & \Checkmark  & 61.3  & 67.3 & 62.4 &63.7      \\
     \rowcolor{gray!15}
   \Checkmark  & \Checkmark   & \textbf{66.6}  & \textbf{68.0} & \textbf{67.2} & \textbf{67.3}    \\
     \bottomrule
    \end{tabular}
   }
  \vspace{-3mm}
  \label{tab:Fusion}

\end{table}
\begin{table}[t]
  \centering
  \vspace{0.12cm}
  \renewcommand\arraystretch{1.05}
  \renewcommand\tabcolsep{5pt}
  \footnotesize
  \caption{Performance Comparison of Different Scanning Strategies for BEV and Instance Features.}
  \vspace{-2mm}
  \resizebox{\columnwidth}{!}{%
    \begin{tabular}{c|c|cccc}
      \toprule
      Module & Scanning Strategy & AP$_{\mathit{ped}}$ & AP$_{\mathit{div}}$ & AP$_{\mathit{bou}}$ & mAP \\
      \midrule
      \multirow{4}{*}{BMF}
        & Single             & 64.3 & 63.7 & 66.6 & 64.9 \\
        & Horizontal         & 65.2 & 66.9 & 66.2 & 66.1 \\
        & Vertical           & 65.4 & 66.8 & 66.7 & 66.3 \\
      \rowcolor{gray!15}
      \cellcolor{white} 
        & Multi-directional  & \textbf{66.6} & \textbf{68.0} & \textbf{67.2} & \textbf{67.3} \\
      \midrule
      \multirow{3}{*}{IMF}
        & Spatial            & 65.2 & 68.0 & 65.9 & 66.4 \\
        & Temporal           & 64.7 & 69.6 & 65.4 & 66.6 \\
      \rowcolor{gray!15}
      \cellcolor{white} 
        & Spatial–temporal   & \textbf{66.6} & \textbf{68.0} & \textbf{67.2} & \textbf{67.3} \\
      \bottomrule
    \end{tabular}%
  }
  \vspace{-6mm}
  \label{tab:Scanning}
\end{table}

\begin{figure*}[t!]
    \centering
    \vspace{0.1cm}
    \setlength{\abovecaptionskip}{-0.1cm}
    \includegraphics[width=1.0\linewidth]{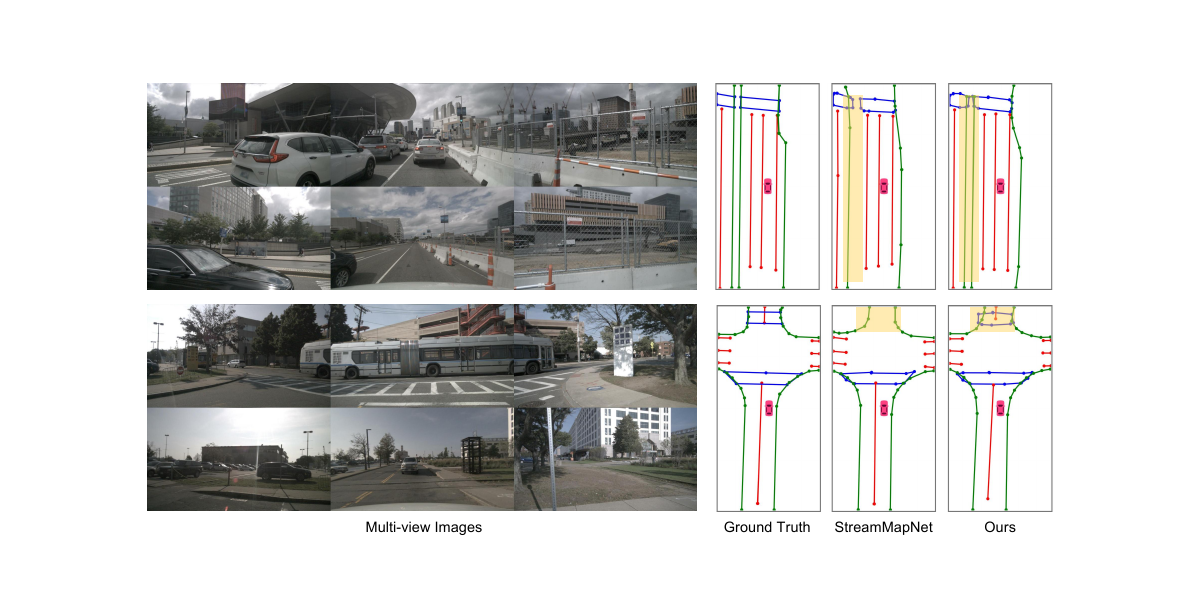}
    \caption{Qualitative visualization comparison with StreamMapNet~\cite{yuan2024streammapnet} across different scenarios. In the HD maps, green lines represent road boundaries, red lines indicate lane dividers, and blue lines denote pedestrian crossings. Best viewed in color.}

    \label{fig:4}
\vspace{-6mm}
\end{figure*}

Furthermore, we evaluate different cost functions for the Hungarian algorithm in the IMF, as shown in Table~\ref{tab:Match}. ``Pts Chamfer Distance'' computes the matching cost based on the Chamfer Distance between points regressed by the regression head, achieving an mAP of 66.3. ``Query Cosine Similarity'' calculates the cost using the cosine similarity between queries, slightly improving the mAP to 65.6. Our chosen method, ``Query $L_2$ Distance'', measures the $L_2$ distance between queries and achieves the highest mAP of 67.3, demonstrating its superiority for instance matching.

{
\begin{table}[t]
\vspace{0.12cm}
\centering

\caption{Ablation Study of Instance Query Matching Methods.}
   \vspace{-2mm}
\resizebox{\columnwidth}{!}{%
 \renewcommand\arraystretch{1.05}
  \renewcommand\tabcolsep{5pt}
  \footnotesize
\begin{tabular}{c|cccc}
\toprule
              Matching Method  & AP$_{\textit{ped}}$ & AP$_{\textit{div}}$ & AP$_{\textit{bou}}$ &  mAP \\ 
\midrule
Pts Chamfer Distance  & 65.1&67.0&64.1&65.4\\
Query Cosine Similarity & 65.0  &  67.1   &  66.9 &66.3    \\
\rowcolor{gray!15}
Query $L_2$ Distance &\textbf{66.6} & \textbf{68.0} & \textbf{67.2} & \textbf{67.3} \\ 

\bottomrule
\end{tabular}%
}
 \vspace{-4mm}
\label{tab:Match}
\end{table}
}

\begin{table}[t]
\vspace{0.2cm}
\centering

\caption{Performance Comparison of SSM Variants in the Gated State Space Block.}
    \vspace{-2mm}
\resizebox{\columnwidth}{!}{%
 \renewcommand\arraystretch{1.05}
  \renewcommand\tabcolsep{7pt}
  \footnotesize
\begin{tabular}{c|cccc}
\toprule
              State Space Model  & AP$_{\textit{ped}}$ & AP$_{\textit{div}}$ & AP$_{\textit{bou}}$ &  mAP \\ 
\midrule
S4~\cite{gu2021efficiently}  &65.1 & 65.0&63.7&64.6\\
   
\rowcolor{gray!15}
DSS~\cite{gupta2022diagonal}  &  \textbf{66.6} & \textbf{68.0} & \textbf{67.2} & \textbf{67.3} \\ 
 Mamba~\cite{gu2023mamba} &  64.1& 66.7&65.1 &65.3
\\

 Mamba-2~\cite{dao2024transformers} & 65.8  &65.4 &64.5 &65.2\\
\bottomrule
\end{tabular}%
}
\vspace{-5mm}

\label{tab:SSM}
\end{table}
Additionally, we compare different SSM variants in the Gated State Space block in Table~\ref{tab:SSM}. DSS~\cite{gupta2022diagonal} achieves the highest mAP of 67.3, outperforming S4~\cite{gu2021efficiently} (64.6 mAP), Mamba~\cite{gu2023mamba} (65.3 mAP), and Mamba-2~\cite{dao2024transformers} (65.2 mAP). These results highlight the importance of choosing appropriate SSM variants. While our framework is conceptually inspired by Mamba~\cite{gu2023mamba}, our implementation adopts DSS~\cite{gupta2022diagonal} due to its superior performance in this task.

We study the impact of Memory Bank size by varying $N$. As shown in Table~\ref{tab:N}, the mAP improves as $N$ increases from 1 to 4. However, the performance drops slightly when $N$ is further increased to 6. This suggests that larger Memory Bank generally enhances performance by incorporating more historical information, while excessively large sizes may introduce redundancy and noise.

\subsection{Qualitative Results}
Fig.~\ref{fig:4} presents qualitative comparisons between StreamMapNet~\cite{yuan2024streammapnet} and MambaMap across several driving scenarios. The results demonstrate the superior performance of MambaMap, particularly in complex or occluded environments. By leveraging its advanced temporal fusion mechanism, MambaMap effectively integrates multi-frame information to model road elements with higher precision. This ability to capture long-range temporal dependencies significantly enhances its robustness and reliability in real-world driving conditions.

\begin{table}[t]

  \caption{Ablation study of the Memory Bank size.}
     \vspace{-2mm}
  \centering
  \renewcommand\arraystretch{1.05}
  \renewcommand\tabcolsep{7pt}
  \footnotesize
   \resizebox{\columnwidth}{!}{
  \begin{tabular}{c|cccc}
    \toprule
    Memory Bank Size & AP$_{\textit{ped}}$ & AP$_{\textit{div}}$ & AP$_{\textit{bou}}$ &  mAP \\ 
    \midrule
       1 & 65.0& 67.8& 66.9& 66.6    \\
        2&  66.4& 67.9 &67.0 &  67.1  \\ 
       
      \rowcolor{gray!15} 4
        &\textbf{66.6}  & \textbf{68.0} & \textbf{67.2} & \textbf{67.3}    \\
        6& 65.3 & 67.4 & 66.8 &  66.5    \\
     \bottomrule
    \end{tabular}
   }
 \vspace{-6mm}

  \label{tab:N}
 
\end{table}

\section{Conclusion}
\label{sec:conclusion}

In this paper, we introduce MambaMap, a novel framework designed to enhance temporal fusion for online vectorized HD map construction. By leveraging a dynamic memory mechanism and a gated SSM, MambaMap efficiently fuses both BEV and instance-level features across multiple time steps, which is capable of capturing long-range dependencies with minimal computational overhead. Our proposed multi-directional and spatial-temporal scanning strategies further enhance feature extraction and temporal consistency. Experiments on the nuScenes and Argoverse2 datasets demonstrate that our proposed approach achieves the state-of-the-art performance, exhibiting strong robustness and generalization across diverse scenarios. For future work, we plan to extend MambaMap to address other BEV perception tasks, such as 3D object detection and motion prediction, thereby broadening its applicability in robotics.
\section*{Acknowledgments}
This work is supported by National Natural Science Foundation of China under Grant No.62376244.

\addtolength{\textheight}{-0cm}   %

\bibliographystyle{unsrt}
\bibliography{ref}

\end{document}